# Mining Frames: Association Rule Mining as an Unsupervised Framework for News Framing Detection

**Hager H. Mustafa**

*Anmat Computational Analysis Lab*

ORCID: 0009-0003-5075-8442

info@anmat.media | www.anmat.media

## Abstract

Computational framing analysis has been dominated by supervised classification methods requiring large annotated corpora and predefined frame taxonomies, limiting their applicability to non-English languages, novel events, and resource-scarce contexts. This paper presents Association Rule Mining (ARM) as an unsupervised, annotation-free framework for news framing detection, addressed to interdisciplinary venues spanning computational media studies, data journalism, and communication research. The primary validation is the Framing Gaza project: 25,512 English-language news articles from 16 outlets across the United States, United Kingdom, France, and Germany, covering the first year of the Gaza war (October 2023 to October 2024). ARM (Apriori algorithm, mlxtend) is applied to headlines and body text with platform-weighted support calculation (weights of 1/√(n) per outlet), POS-filtered n-gram preprocessing, and lift-based framing distinctiveness scoring. Results demonstrate that ARM inductively surfaces interpretable framing patterns without predefined categories. A platform-level closed-rule analysis across all 16 outlets identifies the West Bank triple cluster ({bank, west, israeli}) as the most structurally robust cross-platform co-occurrence (a marker of topical salience rather than, in itself, a frame), present as a closed rule in all 16 outlets (lift 58.3, confidence 97.8%), and {israeli, strike} to {kill} as the dominant military frame in 14 of 16 outlets (lift 8.7, confidence 56.5%). A temporal analysis of body-text closed rules across 12 consecutive months shows event-driven framing pivots without prior event labeling: the World Central Kitchen strike of April 2024 produces the most tightly bound n-gram cluster of any single-event month (lift 6.43), and geographic frames shift from Khan Younis (January 2024, peak lift 6.67) to West Bank (July 2024) to Lebanon-ceasefire (September 2024) in correspondence with documented operational phases. A systematic headline-body framing divergence is identified: headline framing is editorially constructed, while body text is dominated by institutional and procedural language. As a methodological validation study, the paper additionally reports on the Seven Days project, where the framework was tested under acute conditions using FP-Growth on Arabic-language coverage of the 2026 Iran war (1,575 articles, 5 outlets, 10-day window), demonstrating that replacing Apriori with FP-Growth enables sentence-level transaction analysis with substantially reduced memory overhead, while producing comparable framing patterns. All findings were validated through editorial partnership: two published investigative stories, on Daraj Media and Saheeh Masr, developed by independent editorial teams. Preprocessing code is released openly; the full ARM pipeline is documented but not released to preserve the framework's operational deployability.

## 1. Introduction

Media framing, the selective emphasis of particular aspects of events to promote specific problem definitions, causal interpretations, moral evaluations, and treatment recommendations (Entman, 1993), is one of the most consequential mechanisms through which journalism shapes public understanding. Yet the systematic, large-scale detection of framing patterns across corpora of tens of thousands of articles remains a methodological challenge. Existing computational approaches either require extensive manual annotation (Card et al., 2015), impose predefined frame taxonomies that may not capture the specific framing choices of a given corpus, or detect topics rather than frames (Ali and Hassan, 2022). The result is a field that has produced robust methods for English-language supervised framing classification while leaving unsupervised, inductive framing detection largely unaddressed.

This paper proposes Association Rule Mining (ARM) as a solution. ARM is an unsupervised data mining technique that identifies frequent co-occurrence relationships in transactional datasets, producing if-then rules of the form {item A, item B} → {item C} evaluated against support (frequency), confidence (conditional probability), and lift (excess co-occurrence relative to statistical independence). Applied to text corpora, where each document or sentence is treated as a transaction of lexical items, ARM discovers which words consistently appear together at rates far exceeding chance. We argue and demonstrate that high-lift co-occurrence rules correspond to what Gamson and Modigliani (1989) call condensing symbols, the specific lexical packages that carry framing significance in news text. High lift is a necessary but not sufficient condition for framing significance; the interpretive judgment of whether a rule constitutes a condensing symbol is provided by the journalist-researcher in the human-in-the-loop review step described in Section 3.

The primary empirical contribution is the Framing Gaza project: ARM analysis of 25,512 English-language news articles from 16 outlets across four countries, covering the first year of the Gaza war. This constitutes, to our knowledge, the largest computational framing study of Gaza war coverage to date. The paper additionally reports on the Seven Days project as a methodological validation case, in which the ARM framework was adapted for Arabic-language coverage under acute news conditions using FP-Growth, demonstrating the framework's scalability and its behavior under a different algorithm, language, and corpus scale.

### 1.1 Research Questions

**RQ1:** Can ARM, applied as an unsupervised text mining method, surface interpretable media framing patterns from large-scale news corpora without predefined frame categories or annotated training data?

**RQ2:** What framing patterns characterize English-language Global North coverage of the Gaza war, and how do these patterns differ across outlet tiers, time periods, and text levels (headlines versus body text)?

RQ3: Does substituting Apriori with FP-Growth and document-level with sentence-level transactions produce more specific framing detection, and what are the computational trade-offs?

**RQ4:** How does ARM-based framing detection compare to existing computational framing analysis methods in terms of annotation requirements, output interpretability, and journalistic utility?

### 1.2 Contributions

The paper makes four contributions. First, it introduces ARM-based framing detection as a formally described, replicable methodology with documented parameter choices. Second, it presents the Framing Gaza dataset, 25,512 articles spanning one year across 16 outlets, as the primary empirical evidence, supplemented by the Seven Days methodological validation. Third, it identifies and quantifies a systematic headline-body framing divergence as an independent empirical finding. Fourth, it provides the first documented case of a complete pipeline from ARM computation to published investigative story in both English-language international and Arabic-language regional journalism contexts. We additionally propose editorial uptake - whether a pattern can be developed into a published, fact-checked story by an independent editorial team - as a validity standard for journalism-facing framing research, complementary to inter-annotator reliability.

### 1.3 Code Disclosure

The preprocessing pipeline, including the English tokenization, POS filtering, and lemmatization functions, and the Arabic normalization and multi-word expression merging functions, is released openly to support reproducibility and enable adaptation by researchers and journalists. The full ARM pipeline, including threshold configuration, weighting strategy, topic labeling dictionaries, and rule filtering logic, is documented in this paper but not released as executable code. This decision reflects the operational context of the Anmat Lab: the framework's value for journalism partners lies not only in the algorithmic steps, which are straightforward to implement from this paper's documentation, but in the domain expertise, source relationships, and interpretive judgment that the full pipeline embodies. Releasing the preprocessing code enables methodological reproducibility; retaining the full pipeline preserves the operational deployability of the framework as a journalism tool.

## 2. Literature Review

This section reviews framing theory foundations, computational framing analysis approaches, ARM's theoretical and empirical basis for framing detection, and the Arabic NLP context relevant to the Seven Days validation.

### 2.1 Framing Theory

Entman's (1993) foundational definition identifies four framing functions: problem definition, causal interpretation, moral evaluation, and treatment recommendation. These functions manifest in text through lexical choices that selectively emphasize particular aspects of reality. Gamson and Modigliani (1989) identify condensing symbols, catchphrases, exemplars, metaphors, and visual images, as the textual devices through which frames are carried. Pan and Kosicki (1993) locate frames in the syntactical, script, thematic, and rhetorical structure of news text. Semetko and Valkenburg (2000) operationalize five generic news frames, conflict, attribution of responsibility, human interest, economic consequences, and morality, that have become the dominant empirical taxonomy for news framing research.

The theoretical account of why ARM detects frames follows from Gamson and Modigliani (1989). Condensing symbols are by definition items that consistently co-occur: a catchphrase packages multiple frame elements in a single expression, and exemplars recur across articles covering similar events. High-lift ARM rules identify pairs and clusters of items that co-occur far more frequently than independence would predict, precisely the statistical signature of condensing symbol clusters. ARM's lift metric is therefore a measure of condensing symbol strength, subject to the interpretive constraint that high lift is necessary but not sufficient for framing significance.

### 2.2 Computational Framing Analysis

Ali and Hassan (2022) provide the most comprehensive survey of computational framing analysis (CFA), identifying supervised classification and unsupervised topic modeling as the two dominant paradigms. The Media Frames Corpus (Card et al., 2015), annotated across 15 generic framing dimensions, has become the primary benchmark. Otmakhova et al. (2024) identify the fundamental limitation of unsupervised approaches: they detect what is discussed, not how it is framed. Ali and Hassan (2023) introduce SUFA, a dependency parsing approach that captures semantic relationships rather than bag-of-words co-occurrence. Jiang et al. (2025) use community detection in framing element networks as an alternative unsupervised approach.

Gaza war coverage has attracted specific computational attention. The FIGNEWS 2024 shared task (Zaghouani et al., 2024) annotated bias and framing in multilingual Gaza coverage. Kaur and Arora (2025) apply Galtung's war/peace journalism framework to the same conflict using LLMs and frame semantics on a corpus spanning October 2023 to February 2024, finding higher war-based reporting and substantial differences across US, UK, and Middle Eastern outlets in how assailant and victim roles are framed. Pastorino, Sivakumar, and Moosavi (2024) evaluate GPT-4, GPT-3.5, and FLAN-T5 on framing bias detection, finding that LLMs can identify generic frames but produce inconsistent outputs when frame distinctions require nuanced contextual judgment. The present work differs from all three in that it applies an annotation-free inductive method with no predefined frame taxonomy, covers the full first year of the conflict across 16 outlets, and connects computational outputs directly to published investigative stories.

### 2.3 ARM in Text Analysis

Agrawal, Imielinski, and Swami (1993) formalized ARM, and Agrawal and Srikant (1994) introduced Apriori. Han, Pei, and Yin (2000) developed FP-Growth, which replaces Apriori's candidate generation with a compressed FP-tree, achieving time complexity of $O(|D| \times l)$ versus Apriori's worst-case $O(2^{|I|})$. Empirical benchmarks consistently show FP-Growth to be an order of magnitude faster at equivalent support thresholds (Han et al., 2000). Springer et al. (2022) survey ARM applications in social media text for topic detection, but no prior work applies ARM as a framing detection method in news corpora. The ACL 2020 script induction work that modeled narrative chains as ARM problems provides conceptual support for ARM's narrative relevance but was not applied to news framing.

### 2.4 Arabic NLP Context

Arabic NLP faces challenges from morphological complexity, orthographic ambiguity, and dialectal variation. AraBERT (Antoun et al., 2020) and its successors (Abdul-Mageed et al., 2021; Inoue et al., 2021) have advanced Arabic NLP for supervised tasks but require labeled data. ARM-based approaches for Arabic news framing, as demonstrated in the Seven Days project, require only tokenization and normalization, making them applicable without language-specific pre-trained models.

# 3. Methodology

## 3.1 Framework Overview

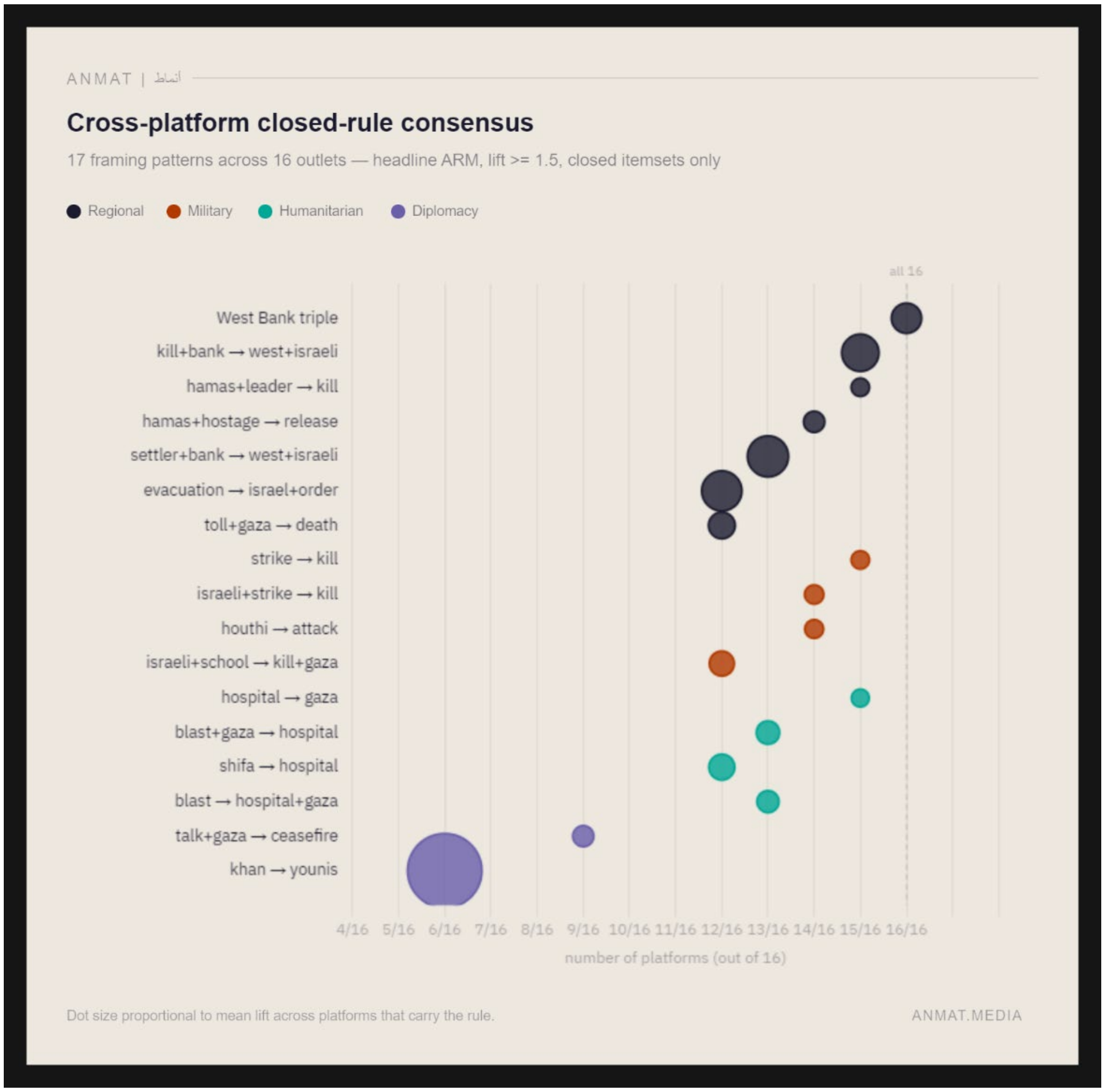

*Figure 1. Closed versus non-closed itemsets: definition and Framing Gaza examples.*

The Anmat ARM framework treats news articles as transactional databases. Each document (or sentence, depending on the project) is encoded as a basket of lexical items. ARM algorithms identify which items consistently co-occur above minimum frequency thresholds. Rules are filtered by lift to retain statistically non-random associations, enriched with topic labels and strength categories, and reviewed by journalist-researchers for journalistic interpretability.

This inductive approach differs from supervised CFA: no predefined frame taxonomy is imposed on the data, no labeled training corpus is required, and the framework discovers whatever co-occurrence structures exist in the corpus. The trade-off is interpretive dependence on human expertise: ARM rules identify that certain words consistently co-occur, but the framing significance of that co-occurrence must be assessed by a domain-expert analyst.

A central commitment of the framework is that a co-occurrence pattern is not, in itself, evidence of a frame. High lift establishes that a lexical cluster is statistically distinctive; whether that cluster constitutes a frame - in Entman's (1993) sense of selecting and making salient a particular problem definition, causal interpretation, moral evaluation, or treatment recommendation - is an interpretive judgment made in context by the journalist-researcher. We therefore treat a rule as framing-significant only when it satisfies two conditions: (i) high lift (statistical distinctiveness), and (ii) the binding of an evaluative, attributive, or relational element, rather than the mere co-reference of entities or toponyms. Under this criterion, {ceasefire, attack} is framing-significant - it embeds a diplomatic term within a military propositional context, constructing the ceasefire as contested - whereas {bank, west, israeli}, though high-lift, is primarily a marker of topical aboutness (the West Bank as a recurrent location) unless an evaluative element accompanies it. The human-in-the-loop review step is therefore not a remedy for a limitation but a constitutive part of the method: ARM proposes candidate frames, and the analyst, reading them in context, confirms which candidates carry framing significance.

### 3.2 Corpora

#### *3.2.1 Framing Gaza (Primary)*

The corpus comprises 25,512 articles from 16 English-language outlets across four countries: the United States (NYT, Washington Post, Fox News, NY Post, Newsweek, CNN), the United Kingdom (Daily Mail, Guardian, Independent, BBC, Sky News), France (France24), and Germany (DW). Collection used News Please and Media Cloud in combination, reducing dependence on any single source. Outlet selection criteria: English-language publication, high reach per SimilarWeb, and minimum article count (smallest: Reuters, 293 articles). The corpus covers October 7, 2023 to October 7, 2024, the full first year of the war. The resulting dataset is openly available (Anmat, 2025).

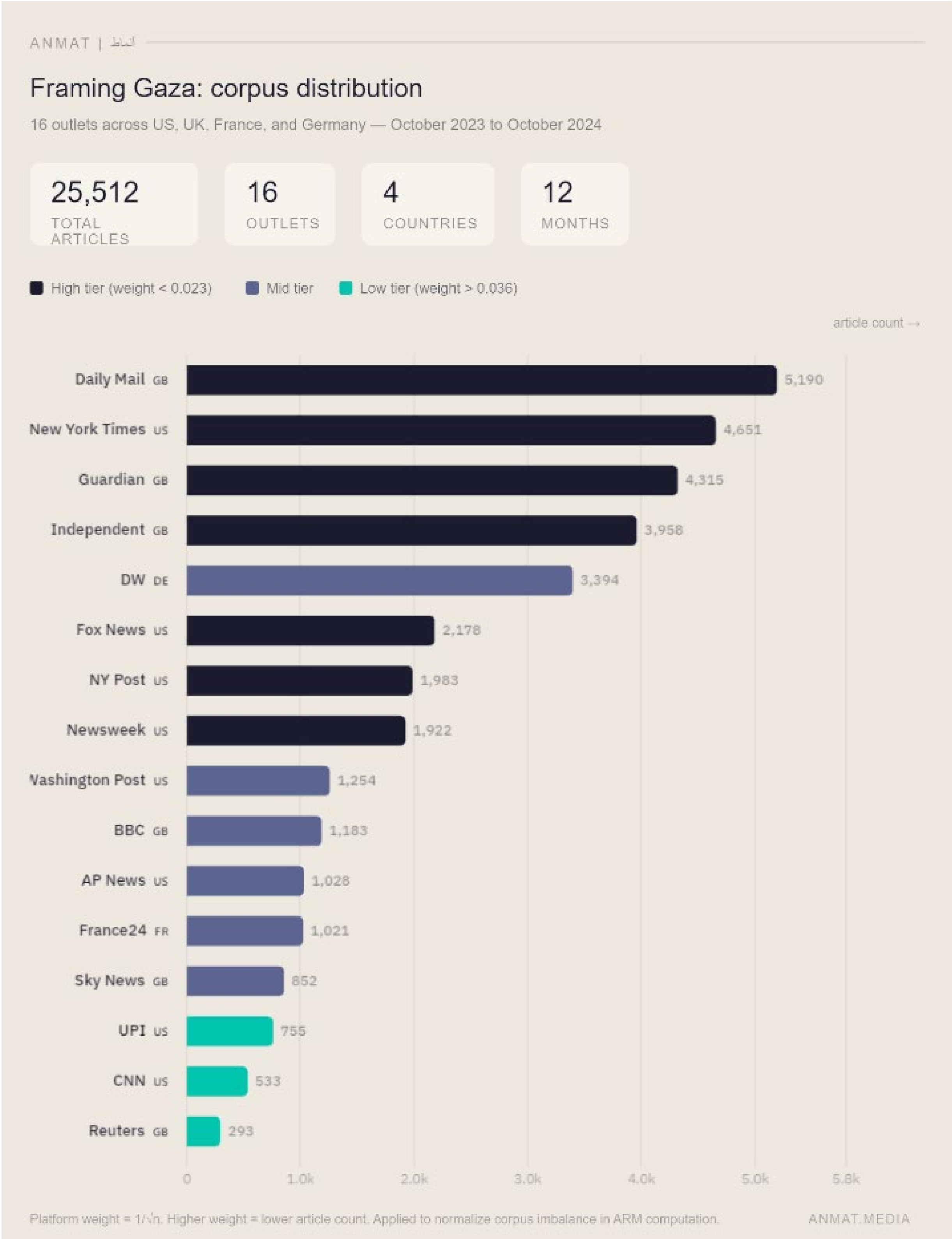

*Figure 2. Framing Gaza corpus: outlet distribution, article counts, platform weights, and tiers.*

### 3.2.2 Seven Days (Methodological Validation)

The Seven Days corpus comprises 1,575 articles from five pan-Arab digital platforms: BBC Arabic, RT Arabic, Al Jazeera, Sky News Arabia, and Al Arabiya. The collection window covers ten days: three days before the American-Israeli strikes on Iran (February 25-27, 2026) and seven days following the onset of hostilities (February 28 to March 6, 2026). Collection used Python open-source scraping libraries. This project serves as a methodological validation case for three reasons: (1) it tests the framework under acute news conditions with a compressed time window; (2) it applies FP-Growth instead of Apriori, enabling algorithm comparison; and (3) it applies sentence-level transaction encoding to body text, addressing the document-level limitation of the Gaza analysis.

## 3.3 Text Preprocessing

### *3.3.1 English Pipeline (Framing Gaza)*

The preprocessing sequence applies to each article: (1) boilerplate removal targeting platform-specific navigation and subscription text; (2) tokenization using NLTK's word_tokenize; (3) POS tagging using spaCy's en_core_web_sm; (4) lemmatization using NLTK's WordNetLemmatizer; (5) stopword removal combining NLTK's English stopword list with a custom 60+ term set covering platform names, generic news vocabulary, and filler expressions; (6) POS-based filtering retaining only tokens in four syntactically meaningful patterns: (NOUN, NOUN), (ADJ, NOUN), (VERB, NOUN), (NOUN, VERB); and (7) n-gram construction generating unigrams, bigrams, and trigrams, with tokens filtered to minimum length four characters and a maximum of two underscore-joined components.

### *3.3.2 Arabic Pipeline (Seven Days)*

Arabic preprocessing applies: (1) a normalization function removing diacritics, normalizing Alef variants, normalizing Alif Maqsura, and removing Tatweel; (2) multi-word expression (MWE) pre-merging using greedy longest-first matching on a curated list of named entities, political titles, and fixed phrases (e.g., الحرس_الثوري_الإيراني, مجلس_الأمن_الدولي); (3) ISRI stemming on non-MWE tokens; (4) Arabic stopword removal combining NLTK's Arabic base set with news-specific additions (attribution verbs, platform names, high-frequency filler expressions); and (5) boilerplate removal targeting Arabic-specific structures. MWE tokens bypass stemming and stopword removal.

### 3.4 Transaction Unit Definition

| Project | Text Type | Transaction Unit | Rationale |
|---|---|---|---|
| Framing Gaza | Headlines | Full headline | Short text; document is natural unit |
| Framing Gaza | Body text | Full article | 25,512-article scale makes sentence-level Apriori computationally infeasible (memory overflow in pilot) |
| Seven Days | Headlines | Full headline | Short text; document is natural unit |
| Seven Days | Body text | Sentence/clause (split on . ، ? !) | Smaller corpus + FP-Growth efficiency enables finer granularity |

*Table 1. Transaction unit definitions.*

The document-level encoding in Gaza body text reflects a computational constraint: an attempt to apply sentence-level Apriori to the 25,512-article corpus produced memory overflow. This is the primary methodological limitation of the Gaza analysis and directly motivates the Seven Days validation, which tests whether FP-Growth enables sentence-level analysis. The outstanding empirical question, whether FP-Growth can handle sentence-level Gaza-scale analysis, requires profiling against the specific hardware constraints of the research environment and is identified as future work.

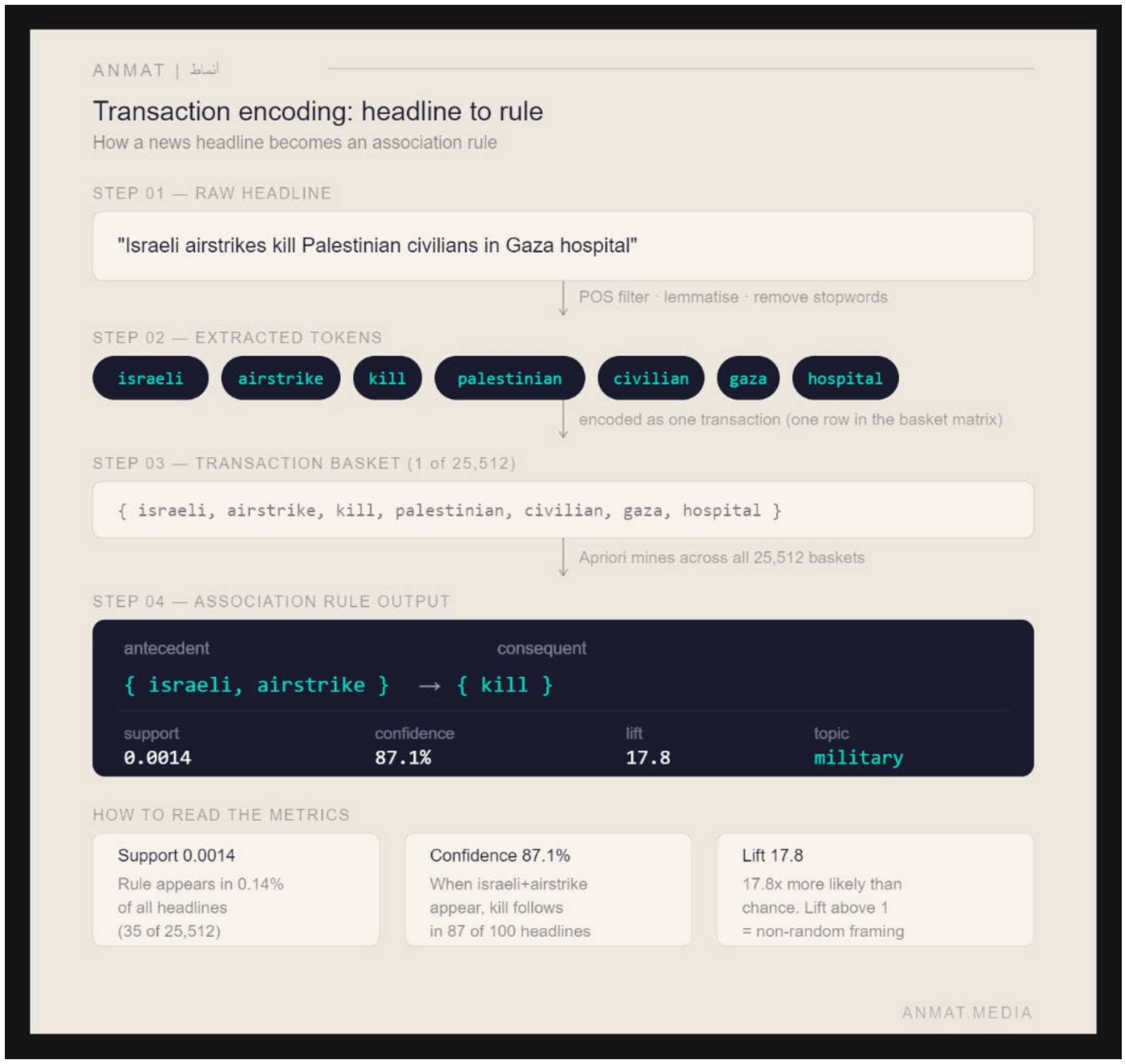

*Figure 3. ARM transaction encoding: from raw headline to association rule.*

### 3.5 Algorithm Selection

Apriori (Agrawal and Srikant, 1994) was selected for the Gaza project as the foundational analysis. Its iterative candidate generation logic is more transparent than FP-Growth's tree traversal, which is relevant for a journalism-facing framework where methodological decisions must be explainable to non-technical partners. FP-Growth (Han, Pei, and Yin, 2000) was adopted for Seven Days because sentence-level transaction encoding at 1,575-article scale required substantially lower memory overhead. Both algorithms are implemented via mlxtend (Raschka, 2018) and produce identical frequent itemsets and association rules for equivalent support thresholds. Their outputs are therefore directly comparable; the choice between them is determined by computational performance rather than by analytical differences.

### 3.6 Threshold Configuration

| Project | Text | Level | Min Support | Min Confidence | Lift Filter |
|---|---|---|---|---|---|
| Gaza | Headlines | General (weighted) | 0.1% | 10% | > 1 |
| Gaza | Headlines | Per-platform | 0.1% | 10% | > 1 |
| Gaza | Body text | General (weighted) | 0.7% (weighted count) | 70% | > 1 |
| Gaza | Body text | Per-platform | 0.7% (weighted count) | 70% | > 1 |
| Seven Days | Headlines | General (weighted) | 0.5% | 5% | > 1 |
| Seven Days | Body text | General (weighted) | 0.5% | 5% | > 1 |

*Table 2. ARM threshold values.*

The 14-fold difference in minimum confidence between Gaza body text (70%) and Seven Days (5%) reflects the different transaction granularities. Document-level transactions are semantically diffuse: a 1,000-word article can contain many topics, so high confidence is needed to filter rules that reflect only accidental co-occurrence within a long document. Sentence-level transactions are semantically tight: items in the same sentence almost certainly share a propositional relationship, so lower confidence thresholds are appropriate. A formal sensitivity analysis varying thresholds is identified as future work.

Platform weighting applies $1/\sqrt{(n)}$ per outlet, where n is the outlet's article count. This reduces the dominance of the Daily Mail (5,190 articles, weight 0.0139) relative to Reuters (293 articles, weight 0.0584) without fully equalizing their contributions. The inverse square-root formula was chosen for its simplicity and interpretability; its theoretical justification relative to alternatives (inverse frequency, equal weighting) warrants future systematic comparison.

### 3.7 Rule Filtering and Enrichment

Post-processing filters remove redundant rules where the antecedent is a subset of the consequent with identical support (non-closed itemsets). Topic labeling assigns each rule to one of five categories based on keyword matching: military, humanitarian, diplomacy, geopolitics, and regional. Rules matching no category are labeled Other. Lift categorization assigns qualitative strength labels: Weak (lift < 2), Moderate (2 ≤ lift < 5), Strong (5 ≤ lift < 10), Very Strong (lift ≥ 10).

### 3.8 Journalistic Validation Pipeline

ARM outputs (structured CSV files with support, confidence, lift, topic, and rule strength) are provided to independent editorial partners who review rule clusters and develop publishable stories. The separation between the ARM analysis team (Anmat) and the editorial writing team (Daraj Media and Saheeh Masr) is methodologically significant: editorial partners are not involved in ARM parameter choices or rule interpretation, and their decision to build and publish stories on the ARM outputs constitutes an independent judgment of journalistic significance. This pipeline constitutes the primary validation of the framework for journalism-facing research contexts. A formal inter-annotator reliability study, which would address the reproducibility requirements of NLP venues, is identified as future work (Section 5.7).

We adopt editorial uptake as the framework's validation standard. A candidate rule is considered validated when an editorial team, working independently of the ARM analysis, develops it into a published, fact-checked story. This establishes the ecological validity and journalistic usability of the detected patterns: it tests whether a computationally surfaced pattern can sustain substantive journalism under real editorial scrutiny. Editorial-uptake validation is a measure of validity (is the pattern meaningful and usable?), complementary to, and distinct from, inter-annotator reliability (would independent coders agree on a label?), which measures the reproducibility of an interpretation scheme and which we leave to future work (Section 5.7). We propose editorial uptake as a complementary validation standard for journalism-facing computational framing research - more externally demanding than annotation-benchmark agreement - while noting its present limits in this study: a small number of validated stories, and a validation process that is expert-driven rather than blind.

| Story | Framing | ARM rule(s) (lift) | Entman function | Source |
|---|---|---|---|---|
| Daraj Media, "How Is Language Used to Objectify Palestinians?" (2025) | Objectification of Palestinians; Israeli agency; law-enforcement lens | {palestinian, strike} -> {kill, israeli} (19.7); {police, palestinian} -> {arrest} (35.3) | Causal interpretation; moral evaluation | daraj.media (English) |
| Saheeh Masr (Facebook, 2024), with Anmat | Dominance of Israeli official sources; Hamas framed as a terrorist group | {kill, group} -> {terror, hamas} (52.9); Hamas Health Ministry as sole recurrent Palestinian source in body text (6.0-6.3); per-outlet Israeli-source lift 3-6x | Attribution of responsibility; moral evaluation | facebook.com/SaheehMasr |

*Table 3. Rule provenance: published editorial outputs traced to specific ARM findings.*

A third, related Anmat story - Aljadah Media, "The Iran File: a political monopoly of the narrative" (2025) - applies the initiative's source-attribution approach to a separate corpus of Iranian coverage using an opening-sentence voice classification rather than ARM lift rules. It is noted here as related work demonstrating the portability of the initiative's framing-analysis practice, and is not counted as validation of the present paper's ARM findings.

## 4. Results

### 4.1 Framing Gaza: Headline Analysis

#### *4.1.1 Scale and Rule Distribution*

The Gaza headline analysis produced 3,189 association rules across three outlet tiers (high: 2,716 rules; mid: 381; low: 92). All rules have lift > 1 by construction. Lift values range from 1.0 to 130.5. The high-tier dominance in rule count reflects both the larger number of high-tier outlets (7 of 16) and their higher article volumes.

#### *4.1.2 Framing Patterns*

Three headline framing patterns are analytically significant.

First, the kill-terror-group framing cluster. The rule {kill, group} → {terror, Hamas}, lift 52.9, confidence 82%, support 0.15%, from high-tier outlets, is the highest-lift substantive framing rule in the headline analysis. This rule identifies a condensing symbol cluster in which lethal violence by a group is systematically framed as terrorist action, with Hamas as the named referent. The rule's restriction to the high-tier confirms that this framing is specific to outlets with high editorial selectivity, not a generic feature of Gaza coverage.

Second, Palestinian-civilian-kill co-framing. The rule {Palestinian, strike} → {kill, Israeli}, lift 19.7 in mid-tier and 16.2 in high-tier, support 0.26-0.15%, shows that Palestinian consistently co-occurs with strike and kill, with Israeli as the associated agent. This directional structure, Palestinian as the affected party, Israeli as the agent, defines a causal interpretation frame (Entman, 1993) that assigns agency to Israeli action.

Third, police-arrest framing of Palestinian protesters. The rule {police, Palestinian} → {arrest}, lift 35.3, confidence 35%, identifies a high-tier framing pattern associating Palestinian with police action (arrest) rather than with political agency or civilian status. In the London context ({London, Palestinian} → {police}, lift 33.6), this pattern captures coverage of Palestinian solidarity protests in the UK framed primarily through a law enforcement lens.

| Rule | Support | Confidence | Lift | Tier | Topic |
|---|---|---|---|---|---|
| {kill, group} → {terror, Hamas} | 0.15% | 82% | 52.9 | High | regional |
| {kill, terror} → {group} | 0.16% | 89% | 44.3 | High | Other |
| {Palestinian, Middle} → {East} | 0.14% | 100% | 47.5 | High | Other |
| {police, Palestinian} → {arrest} | 0.11% | 35% | 35.3 | High | Other |
| {London, Palestinian} → {police} | 0.12% | 45% | 34.5 | High | Other |
| {Palestinian, strike} → {kill, Israeli} | 0.26% | 73% | 19.7 | Mid | military |
| {kill} → {Israeli} | 5.20% | 52% | 2.35 | High | regional |
| {Israel} → {Hamas} | 21.6% | 43% | 1.42 | High | regional |

*Table 4. Selected Framing Gaza headline association rules. Support values are outlet-weighted.*

### *4.1.3 Platform-Level Closed Rule Analysis*

This section presents a post-hoc closed-rule analysis applied to the per-platform pipeline outputs; closure filtering was not part of the original Framing Gaza pipeline and is reported here as an independent methodological step. The analysis computes closed association rules for each of the 16 outlets individually, applying a closure filter that retains only rules whose antecedent-consequent union has no proper superset with equal support in the same platform output. The filter is applied after topic filtering (topic not equal to Other) and a lift threshold of 1.5, calibrated through workshops with journalist-researchers as the minimum value at which rules carry interpretable framing significance above background co-occurrence. The closure filter reduced raw headline rules by 32 to 68 percent per platform (NYT: 8,232 closed rules from 19,724 total; BBC: 10,993 closed rules). What remains in each platform output are the maximally specific framing structures at each support level: rules where adding any further item would reduce support, confirming that the itemset is genuinely bounded at that frequency. Cross-platform agreement is the primary analytic criterion: a closed rule present across more outlets represents a more robust framing convention, less attributable to outlet-specific editorial or market factors.

Regional frame (12/12 platforms). The West Bank triple cluster is the single most robust cross-platform closed rule in the headline dataset. The rule {bank} -> {west, israeli} (mean lift 58.3, mean confidence 97.8%) and its directional variants ({israeli, bank} -> {west}, lift 50.8, confidence 99.4%; {israel, bank} -> {west}, lift 49.6) appear in all 16 analyzed outlets as closed rules. The rule {west, bank} -> {israeli} (lift 3.49) also reaches 16/16 consensus at lower lift, reflecting higher antecedent support. Because these rules are closed, no superset of the itemset {bank, west, israeli} carries identical support in any of the 16 platform outputs, confirming that the three-item combination is the maximally specific structure the algorithm recovers at this frequency. The West Bank is not merely frequent in Gaza war coverage; it is structurally bound as a three-way association across outlets that differ substantially in political orientation, national origin, and article volume.

Military frame (10 to 11/12 platforms). The rule {israeli, strike} -> {kill} is a closed rule in 14 of 16 platforms (mean lift 8.7, mean confidence 56.5%). Because it is closed, {israeli, strike, kill} has no superset at identical support in these 14 outlets, meaning the three-term combination is the most specific military framing structure the algorithm recovers at this frequency. The broader rule {strike} -> {kill} reaches 15/16 platforms (lift 4.8), while the tighter {israeli, airstrike} -> {kill} (6 platforms, lift 17.8, confidence 87.1%) represents the highest-lift closed military rule in the dataset. The {airstrike} to {kill, israeli} pair (6 platforms, lift 30.9) indicates that the term "airstrike" is a particularly strong predictor of lethal framing when it appears in headlines. Iran-related rules reach 12/16 platforms: {israel, attack} -> {iran} (lift 5.1) and {israel, iran} -> {attack} (lift 5.1). The Houthi attack cluster appears in 14/16 platforms at lift 6.6, extending the military frame to the Red Sea theater.

Humanitarian frame (9 to 11/12 platforms). The rule {hospital} -> {gaza} is a closed rule in 15/16 platforms (lift 2.55, confidence 75.1%). The blast-hospital pairing is more specific: {blast, gaza} -> {hospital} (13/16 platforms, lift 27.4, confidence 71.8%) and {shifa} -> {hospital} (12/16 platforms, lift 38.8, confidence 94.3%) identify Al-Shifa hospital as the primary named healthcare institution in the humanitarian framing cluster. The Shifa-hospital rule carries 93.6% confidence: when "shifa" appears in a headline across the 12-platform consensus set, "hospital" co-occurs in 94 out of 100 such headlines. The civilian-kill association does not appear as a closed rule at the 14/16 threshold or above, indicating that civilian framing in headlines is platform-specific rather than a structural consensus.

Diplomacy frame (9/12 platforms). The ceasefire cluster appears as closed rules in 9 platforms: {talk, gaza} -> {ceasefire} (lift 17.1, confidence 54.4%) and {talk} -> {ceasefire, gaza} (lift 14.7), each present in 9 of 16 platforms. The Khan Younis toponym, also identified in the body-text temporal analysis, appears in headline-level closed rules in 6 platforms (lift 257.1, confidence 88.8% and 100% in the two directions), confirming that the geographic name is a maximally bound unit in outlet outputs where it appears.

The platform-level closed-rule analysis covers all 16 outlets in the Framing Gaza corpus.

## 4.2 Framing Gaza: Body Text Analysis

### *4.2.1 Scale and Rule Distribution*

The Gaza body text analysis produced 212,396 rules across the full corpus. Topic distribution: regional (100,142 rules, 47%), Other (65,420, 31%), military (28,708, 14%), diplomacy (10,097, 5%), humanitarian (6,267, 3%), geopolitics (1,762, 1%). Lift values range from 1.0 to 6.3, substantially lower than headline lift values, reflecting the document-level transaction encoding: with longer texts, more items co-occur by chance, compressing the lift range.

### *4.2.2 Institutional Actor Framing*

The most analytically significant body-text finding is the dominance of the Hamas Health Ministry as an institutional framing node. The rule {hamas_health} → {hamas_health_ministry}, support 16.1%, lift 6.0-6.3, and its multiple directional variants, indicate that the Hamas Health

Ministry is the central institutional actor in body-text co-occurrence networks, appearing in 16% of all articles. Its consistent co-occurrence with kill ({kill, hamas_health} → {hamas_health_ministry}, lift 6.0) and with Israeli ({Israeli, hamas_health} → {hamas_health_ministry}, lift 6.0) identifies a framing structure in which casualty counts are sourced from and attributed to this specific institutional entity. This is a causal interpretation frame: the sourcing convention defines who counts the dead and how that count is attributed.

#### *4.2.3 The Ceasefire-Attack Framing Cluster*

The rule {cease_fire, attack} → {cease}, support 17.0%, lift 4.68, and its inverse {cease, attack} → {cease_fire}, lift 4.68, identify a body-text framing pattern in which ceasefire and attack consistently co-occur. This is not a coincidental co-occurrence: lift 4.68 indicates that these terms appear together nearly five times more often than independence would predict. The framing significance is that the diplomatic vocabulary of ceasefire is systematically embedded in the same propositional context as military vocabulary (attack), constructing ceasefire as a contested or conditional term rather than an unambiguous political goal.

#### *4.2.4 Temporal Framing Trajectory*

To track how framing structures evolve across the coverage period, closed association rules are computed on body text segmented by publication month across the full 13-month window (October 2023 to October 2024). The closure filter is applied after topic filtering and the body-text thresholds established in section 3.6 (support >= 0.150, confidence >= 0.700). Closed-rule counts per month range from 1,770 (November 2023) to 6,848 (September 2024) across the 12 comparable months. October 2024 is excluded from cross-month comparison: its raw rule volume is 715,661, compared to a range of 36,000 to 131,000 for all other months. This ratio is consistent with anniversary-period article accumulation in the corpus rather than a framing shift, and its closed-rule count (89,001) is an order of magnitude above any other month. Its findings are reported separately below.

A methodological note on transaction units. The body text pipeline generates n-gram tokens (multi-word expressions such as israeli_prime_minister, khan_younis, gaza_health_ministry, world_central_kitchen) rather than the single-word tokens used in headline analysis. These n-grams are bound during preprocessing and function as atomic units in ARM transactions. Their dominance in the top closed rules of each month confirms that named entities and institutional actors, not generic political vocabulary, are the primary structural units of body-text framing in this corpus.

October and November 2023 (war onset). The highest-lift closed rules in both months center on the Netanyahu n-gram cluster: {israeli_prime} -> {israeli_prime_minister, prime_minister} (October, lift 6.32) and {israeli_prime_minister, hamas} <-> {israeli_prime} (November, lift 6.53, confidence 100% in the Hamas-to-cluster direction). The Israeli Prime Minister and Hamas are bound as a paired unit from the war's first weeks, establishing a bilateral principal frame that names leadership rather than civilian populations. November's explicit binding of Hamas within the Netanyahu cluster at lift 6.53 indicates that the leadership pair tightened further as the war advanced into its second month.

December 2023 and January 2024 (Khan Younis operations). The Khan Younis n-gram cluster becomes dominant in December (lift 5.98) and reaches its peak in January 2024 (lift 6.67), the highest lift value across any month for any pattern in the 12-month comparable window. The rule {khan_younis} <-> {khan, younis} (confidence 100% in both directions) and {younis} -> {khan_younis, gaza} (confidence 99.2%) establish Khan Younis as a tightly bound geographic frame coinciding with the IDF ground advance in southern Gaza. The algorithm surfaces this without geographic tagging or event coding: the closure filter identifies {khan_younis, khan, younis, gaza} as the maximally specific itemset at its support level.

February 2024 (ceasefire negotiations). The ceasefire n-gram cluster (cease_fire, cease) appears for the first time as the dominant top closed rules (lift 5.48, confidence 99.6%). The rule {cease_fire} -> {israel, cease} (lift 5.48) and {cease} -> {israel, cease_fire} (lift 5.46) indicate that ceasefire discourse in body text is structurally bound to Israel as a named party. Netanyahu co-occurs as a secondary cluster in the same month.

March 2024 (casualty documentation). The Gaza Health Ministry n-gram cluster (gaza_health_ministry) becomes dominant (lift 5.95). The rule {gaza_health_ministry} <-> {health_ministry, gaza_health} (confidence 100% and 97.4%) binds the institution's name as a stable n-gram unit across articles. This coincides with the period when the Ministry's cumulative casualty figures became the primary evidentiary source for international reporting. The closed rule confirms that the Ministry functions as a structurally embedded sourcing institution in body text, not merely an incidental reference.

April 2024 (World Central Kitchen strike). The World Central Kitchen cluster dominates with lift 6.43 and confidence 97.8%. The rule {israel, central_kitchen} -> {world_central_kitchen} (lift 6.43, confidence 100%) and {world_central} -> {israel, world_central_kitchen} (confidence 97.8%) establish the organization as a maximally bound n-gram in April body text. The Israeli military strike on a World Central Kitchen convoy on April 1, 2024, which killed seven aid workers, produces the most tightly bound event-specific closed rule of any non-operational month in the dataset. No other single event in the comparable 12-month window generates comparable n-gram binding around a named organization.

May and June 2024 (alternating frames). The Netanyahu cluster re-emerges in May (lift 5.98), reflecting the ICC arrest warrant context and the Rafah offensive. June sees a return of the Gaza Health Ministry cluster (lift 6.14), indicating resumed casualty-count dominance in body text as the Rafah operation generated renewed reporting on Ministry figures.

July 2024 (West Bank escalation). The West Bank n-gram cluster (west_bank) appears as the dominant pattern for the first time (lift 5.06). The rule {west, bank} -> {west_bank} (confidence 99%) and {west_bank} -> {west, bank} (confidence 100%) establish the West Bank as a structurally distinct geographic frame in July body text, reflecting the documented escalation of Israeli military operations in Jenin and other West Bank locations. This is the only month in which a non-Gaza geographic frame dominates the closed rule output.

August 2024 (diplomatic maneuvering). The Netanyahu cluster returns (lift 5.66), reflecting the period following the Haniyeh assassination in Tehran and subsequent hostage-deal negotiations.

September 2024 (Lebanon escalation and ceasefire pressure). The ceasefire cluster resurges as the dominant pattern (lift 5.37). Simultaneously, military-topic closed rules jump to 2,379 (compared to 317 to 667 in the preceding summer months) and geopolitics-topic closed rules reach 619 from near zero in prior months. This dual shift coincides with the Israeli military escalation in Lebanon, including the pager attack and subsequent airstrikes, and renewed international ceasefire pressure. The algorithm surfaces both the diplomatic vocabulary surge and the military rule expansion in the same month without event labeling.

The temporal trajectory demonstrates a methodologically significant property of closed-rule ARM: the algorithm detects framing pivots that correspond to documentable events the analysis has no prior access to. The April 2024 World Central Kitchen finding is the clearest case: a single event produces the most tightly bound humanitarian n-gram cluster of any non-operational month, recoverable from body-text co-occurrence structure alone, without event coding, geographic tagging, or named entity recognition applied as a pre-analysis filter. This confirms that ARM functions as a framing detection method sensitive to temporal shifts in discourse structure, not only to aggregate corpus patterns.

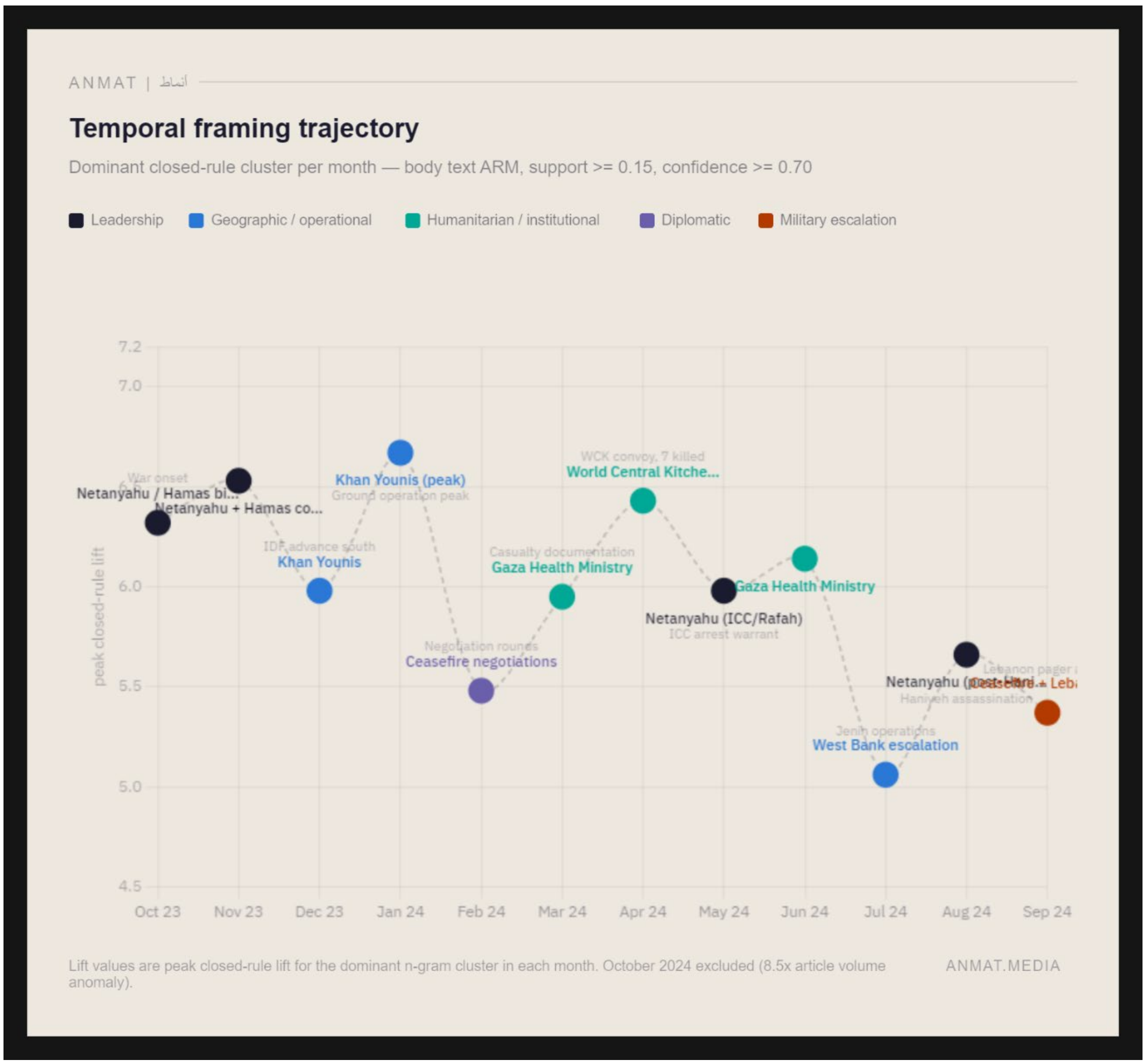

*Figure 4. Temporal framing trajectory: dominant closed-rule cluster per month, October 2023 to September 2024.*

### *4.2.5 Headline-Body Divergence*

A systematic divergence between headline and body-text framing is observable across the corpus. Headlines produce high-lift rules (maximum 130.5) around emotionally charged and politically explicit lexical packages: kill-terror-group, Palestinian-strike-kill. Body text produces lower-lift rules (maximum 6.3) around institutional actors and procedural vocabulary: Hamas Health Ministry, Prime Minister Benjamin Netanyahu, West Bank attack, ceasefire. This divergence is consistent with an editorial structure in which headlines function as framing devices designed to attract reader engagement, while body text reproduces more procedural, source-attributed content. The document-level body-text encoding may partially suppress this divergence by averaging across diverse content within each article; sentence-level encoding (as in the Seven Days validation) would provide a sharper test.

### 4.3 Seven Days: Methodological Validation

#### *4.3.1 FP-Growth versus Apriori: Algorithm Comparison*

The Seven Days project applies FP-Growth with sentence-level transaction encoding. This combination was computationally feasible at 1,575-article scale where Apriori with sentence-level encoding would have required substantially more memory due to candidate generation overhead. The resulting rule set (191 headline rules, 1,781 body-text rules) shows higher lift ceiling values than the Gaza body-text analysis (body-text lift maximum 185.7 in Seven Days versus 6.3 in Gaza), directly attributable to the sentence-level encoding: rules within a single sentence represent tight propositional co-occurrence, producing more distinctive statistical associations.

The rule {defense, interception} → {Saudi Arabia, destruction}, lift 185.7, confidence 98%, is the highest-lift rule in the Seven Days body-text analysis. This rule reflects the Iranian retaliatory strikes on Gulf infrastructure, surfaced at sentence level because the terms appear in the same sentence in military bulletin content. At document level (as in Gaza), this rule would have substantially lower lift because interception and Saudi Arabia would co-occur incidentally across the long body texts of many articles. This comparison indicates that sentence-level encoding produces substantively more specific framing rules than document-level encoding; the change from Apriori to FP-Growth is what makes sentence-level analysis computationally feasible at scale, rather than itself causing the higher specificity. Because the Gaza (document-level, Apriori, English) and Seven Days (sentence-level, FP-Growth, Arabic) analyses differ simultaneously in transaction granularity, algorithm, language, and event, the comparison is confounded; isolating the effect of granularity requires a controlled sentence-level run on the Gaza corpus (Section 5.7).

#### *4.3.2 Arabic Framing Patterns*

The headline analysis produced a tripartite attribution frame as the most analytically significant finding: {war, American} → {Israeli, Iran}, lift 57.7, confidence 52%, indicating that Arabic-language headlines systematically packaged the war as a joint American-Israeli operation against Iran, far more consistently than independence would predict. The rule {Israeli Army} → {announces, Iran}, lift 20.6, confirms that the Israeli Army holds narrative agency as the declaring actor in the conflict's progression. Platform-specific analysis revealed divergent framing of Khamenei's assassination: Gulf outlets (Sky News Arabia, Al Arabiya) produced high-lift rules associating killing and successor and refusal, constructing a succession-crisis frame; RT Arabic produced rules associating Khamenei with Tehran, positioning him as a head of state within the capital rather than a symbol opening a post-regime scenario.

## 5. Discussion

### 5.1 ARM as an Inductive Framing Detector

The results across both projects confirm that ARM surfaces interpretable framing patterns without predefined categories. The kill-terror-group cluster in Gaza headlines, the Hamas Health Ministry sourcing frame in body text, the ceasefire-attack co-framing, and the tripartite attribution frame in Seven Days headlines were all discovered inductively and subsequently validated through journalistic publication. None of these patterns required a predefined codebook entry; all were recognized as journalistically significant by independent editorial teams.

The theoretical account follows from Gamson and Modigliani (1989): frames are carried by condensing symbols that consistently co-occur. ARM's lift metric identifies statistically distinctive co-occurrence, which is the statistical signature of condensing symbol usage. The interpretive constraint, that high lift is necessary but not sufficient for framing significance, is addressed by the human-in-the-loop review step. This is not a weakness unique to ARM but a property of all unsupervised framing methods: topic modeling, community detection, and dependency-parsing approaches all require human interpretation to connect statistical patterns to framing significance.

Several discovered rules map directly onto Entman's (1993) framing functions, established interpretively under the criterion of Section 3.1 rather than asserted from lift alone. The {ceasefire, attack} cluster realizes problem definition (the ceasefire rendered conditional and contested). The {hamas_health_ministry} sourcing cluster, which binds casualty counts to a single named institution, realizes causal interpretation and attribution of responsibility: it encodes who is authorized to count the dead and to whom the count is attributed. The high-lift headline clusters built around kill and terror vocabulary realize moral evaluation. In each case the lift metric locates the candidate; the framing function is confirmed by contextual reading.

### 5.2 The Headline-Body Divergence

The systematic divergence between headline and body-text framing is the most robust cross-project finding. In Gaza, headline lift values reach 130.5 while body-text lift reaches 6.3; in Seven Days, headline lift reaches 57.7 while body-text lift reaches 185.7 (at sentence level). The directional difference, headlines producing more extreme framing clusters, body text more procedural vocabulary, is consistent across both analyses despite the different languages, algorithms, and transaction granularities.

This finding has a consistent interpretation: headline framing is primarily an editorial and commercial act, designed to attract reader engagement through politically salient or emotionally resonant lexical packages. Body text, at document level, reflects a mixture of editorial framing and reproduced institutional content (agency reports, official statements, casualty figures). The sentence-level Seven Days analysis confirms this: high-lift body-text rules reflect tight military and institutional propositions from source documents, not editorial framing choices. The headline-body divergence thus reveals where editorial framing is concentrated in the production of digital news.

### 5.3 Apriori versus FP-Growth: Methodological Implications

The Seven Days validation shows that FP-Growth with sentence-level encoding yields more propositionally specific body-text rules than document-level Apriori. The lift ceiling for body-text rules is 29 times higher at sentence level (185.7 versus 6.3), reflecting the tighter propositional specificity of sentence-level co-occurrence. This comparison speaks to RQ3 with an important caveat: the higher specificity is attributable to sentence-level encoding, not to FP-Growth per se; FP-Growth is the enabling algorithm that makes sentence-level analysis tractable at scale. Because language, algorithm, granularity, and event vary together across the two corpora, the comparison is confounded and cannot isolate the algorithm's contribution. A controlled document-versus-sentence comparison on a single corpus (Section 5.7) is required to make the claim cleanly; we therefore frame FP-Growth with sentence-level transactions as a promising, computationally feasible configuration rather than as a demonstrated analytic superiority.

The outstanding question is whether FP-Growth can handle sentence-level encoding at Gaza corpus scale (25,512 articles). Based on Han et al. (2000), FP-Growth's $O(|D| \times l)$ complexity suggests feasibility in principle, but memory requirements depend on vocabulary size and average sentence length after preprocessing and require empirical profiling. This is the highest-priority future work item, as it would allow direct comparison of sentence-level framing detection across both corpora.

### 5.4 Toward Improved Output Readability: Closed Rules and Future Filtering

A limitation of the current pipeline is the large raw rule volume before filtering, particularly in short-transaction corpora. The Gaza headline analysis produces 3,189 rules at the weighted corpus level; per-platform outputs are substantially larger. As a post-hoc step reported in sections 4.1.3 and 4.2.4, closed itemset filtering was applied to the per-platform and per-month pipeline outputs, reducing raw headline rules by 32 to 68 percent per platform without information loss (Pasquier et al., 1999). This confirms that closure filtering is a viable readability improvement for future pipeline integration. Two further improvements remain as future work: automatic mirror-rule deduplication (retaining only the higher-lift direction for A→B versus B→A pairs) and semantic clustering of related rules into named frame packages using word embedding similarity. Both would reduce the interpretive burden on the journalist-researcher in the rule review step.

### 5.5 Comparison with Alternative CFA Methods

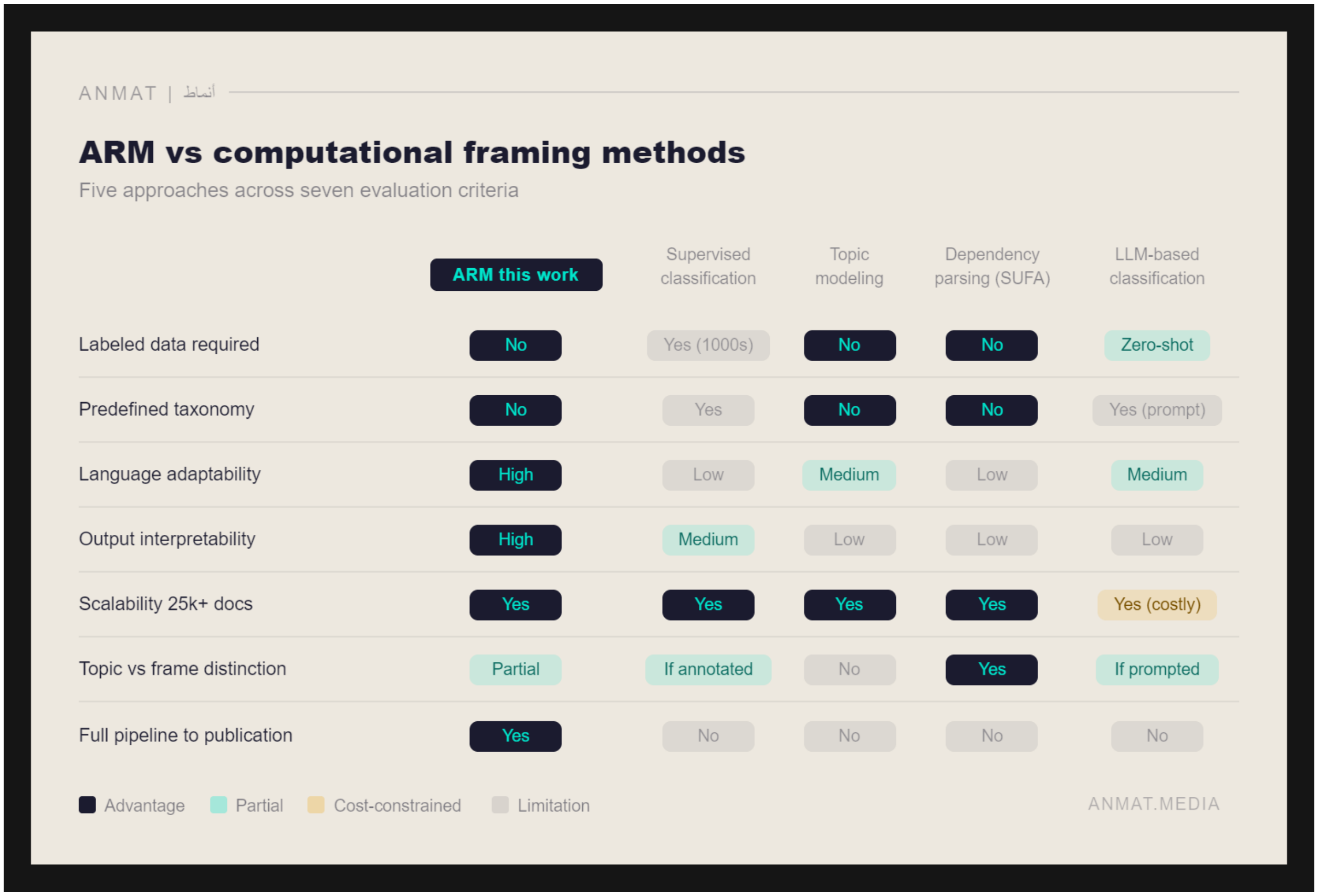

*Figure 5. ARM versus alternative computational framing analysis methods.*

The comparison with LLM-based approaches (Kaur and Arora, 2025; Pastorino et al., 2024) is the most instructive. Kaur and Arora (2025) detect semantic role distinctions that ARM does not produce without human review: their method identifies who is framed as assailant versus victim at the sentence level, which ARM rules do not encode directly. This is a genuine capability gap. However, their approach applies a predefined framework (Galtung's war/peace journalism), covers five months versus twelve in this study, and produces classification labels with no explicit reasoning trace. ARM's lift metric is deterministic and fully reproducible; LLM outputs vary with model version and prompt wording. Pastorino et al. (2024) find that LLMs produce inconsistent frame labels when distinctions require nuanced judgment, a limitation ARM avoids by surfacing only statistically robust co-occurrence patterns above the lift threshold. The practical trade-off is therefore: LLMs offer semantic role detection that ARM lacks; ARM offers full reproducibility, no API dependency, and interpretable if-then rules that non-technical journalists can read directly without model access.

## 5.6 Limitations

### *5.6.1 Interpretive Dependence*

ARM rules identify systematic co-occurrence; the elevation of a co-occurrence to a frame is an interpretive act, performed in context by the journalist-researcher under the two-part criterion of Section 3.1. We treat this interpretive dependence as a design property rather than a deficiency: it is shared by all unsupervised framing methods (topic modeling, community detection, dependency parsing), none of which connects a statistical pattern to framing significance without human judgment. What the framework does not yet provide is an automated criterion for that step; closed-itemset filtering and semantic clustering (Section 5.4) will narrow, but not remove, the interpretive load.

### *5.6.2 Document-Level Body Text in Gaza*

Document-level transaction encoding suppresses sentence-level propositional specificity, producing lower lift values and less specific body-text rules. The ceasefire-attack finding and the Hamas Health Ministry sourcing frame are robust at document level, but finer-grained framing patterns (at the sentence level) may remain undetected in the Gaza analysis.

### *5.6.3 Preprocessing Sensitivity*

ARM outputs are sensitive to stopword list construction and MWE identification. The custom stopword lists in both projects were constructed through iterative corpus inspection, a process that is difficult to fully standardize. Systematic sensitivity analysis varying stopword sets would strengthen methodological reproducibility.

### *5.6.4 Platform Weighting Choice*

The inverse square-root weighting scheme has no theoretical justification relative to alternatives. Different weighting schemes would produce different weighted support values. Future work should compare weighting approaches systematically on the Gaza corpus.

### *5.6.5 Error Analysis*

The present framework documents rules that produced interpretable framing patterns and informed published stories but does not systematically report on rules that were reviewed and discarded. In the Gaza analysis, some high-lift rules in the high-tier reflect platform-specific boilerplate structures (newsletter subscription text, article continuation prompts) that survived preprocessing, rather than genuine framing co-occurrences. Systematic documentation of false positives would strengthen precision claims and inform pipeline improvement.

**5.7 Future Work**

Five research directions follow from this work.

First, sentence-level FP-Growth on the Gaza corpus. This is the highest-priority test of the framework's scalability and would enable direct comparison of sentence-level framing detection across English and Arabic news corpora at equivalent scale.

Second, closed itemset filtering and semantic rule clustering in the pipeline. These improvements (Section 5.4) would substantially reduce raw rule volume and enable automated frame package identification, improving the framework's usability in newsroom contexts.

Third, formal inter-annotator agreement study. Journalist-researcher teams reviewing the same rule set and independently identifying framing patterns would provide a quantitative precision estimate and address the validation requirement of NLP venues.

Fourth, cross-linguistic framing comparison. The Gaza and Seven Days corpora cover overlapping events (the Gaza war appears in both English and Arabic coverage). Controlled ARM analysis across language pairs using equivalent parameters would enable the first ARM-based cross-linguistic framing comparison in the literature.

Fifth, the headline-body divergence as a computable metric. Representing headline and body-text token distributions as term-frequency vectors and computing the cosine distance between them would operationalize the headline-body divergence finding as a corpus-level measurable quantity, enabling platform comparison and longitudinal tracking.

## 6. Conclusion

This paper has demonstrated that Association Rule Mining constitutes a viable, deployable framework for unsupervised framing detection in large-scale news corpora. Applied to 25,512 English-language articles covering the first year of the Gaza war, ARM inductively surfaced interpretable framing patterns, including lexical objectification of Palestinians in high-tier coverage, the Hamas Health Ministry as the dominant institutional actor in casualty sourcing, and ceasefire embedded in military rather than diplomatic framing contexts, without any predefined categories or annotated data. The Seven Days methodological validation demonstrated that FP-Growth with sentence-level encoding yields more propositionally specific body-text rules and is computationally feasible at 1,575-article Arabic corpus scale; because corpus, language, algorithm, and granularity vary together, this is framed as a promising configuration rather than a demonstrated algorithmic superiority, with a controlled sentence-level Gaza-scale analysis identified as the next empirical test.

The headline-body framing divergence, identified quantitatively across both projects and consistent with journalistic editorial practice, represents an independent empirical contribution: headline framing is concentrated, high-lift, and editorially constructed, while body-text framing at document level reflects the procedural vocabulary of institutional sources. ARM's lift metric, which measures excess co-occurrence relative to independence, is the most analytically informative metric for framing analysis, identifying what is distinctive rather than merely frequent.

All findings were validated through editorial partnership: two published investigative stories, on Daraj Media and Saheeh Masr, written by editorial teams independent of the ARM analysis (a third, related Anmat story on Aljadah Media applies the broader approach to Iran coverage and is not counted as validation of the present ARM findings). This validation standard, whether a framing pattern discovered computationally can sustain a published journalistic story, is more demanding and externally meaningful than annotation benchmark agreement, and is proposed as a complementary standard for computational framing analysis research in journalism-facing contexts.

***Statements***

**Data availability**. The Framing Gaza ARM outputs are available as an open dataset on Kaggle (https://www.kaggle.com/datasets/anmatproject/framings-gaza-arm-dataset).
The preprocessing code is released openly; the full ARM pipeline is documented in this paper (Sections 3.3-3.7) but as stated in Section 1.3, is not released.

**Ethics**. All analyzed texts are publicly published news articles; no private or personal data were collected.

**Funding**. This research received no external funding; it was conducted on a voluntary basis.
Competing interests. The author declares no competing interests.

**Use of AI.** All preprocessing, and association-rule-mining analysis code for the Framing Gaza project was written by the author. An AI assistant was used only in preparing this manuscript, for: (i) literature search and reference triage; (ii) methodological review and critique; and (iii) drafting and editing of the text. The author made all methodological and interpretive decisions, verified all reported results and figures against the analysis outputs, and takes full responsibility for the content. The AI is not an author.

**Acknowledgements**. The author thanks the editorial teams at Daraj Media, Aljadah Media, and Saheeh Masr for their collaboration, Public Policy Researcher Kais Attia for his work on Data Collection for Framing Gaza, and Co-Founder and product Manager of Anmat, Ahmed Belal on His work on Data Collection for The Seven Days.

**Published stories cited**. Daraj Media, "How Is Language Used to Objectify Palestinians?", https://daraj.media/en/how-is-language-used-to-objectify-palestinians/; Saheeh Masr (Facebook), https://web.facebook.com/SaheehMasr/posts/pfbid02nudnKKNAq87J9Cpwni5FkVtQ1q9qUodGjiGUntHbAa7oSh2gFBRJkVVXJRg13Pt8l; Aljadah Media, "The Iran File" (related work), https://aljadah.media/archives/84116